\documentclass{sig-alternate}

\usepackage{algorithm}
\usepackage{algorithmic}
\usepackage{graphicx}
\usepackage{subfigure}

\begin{document}
%
% --- Author Metadata here ---
\conferenceinfo{GECCO '10,}{July 7-11,2010, Portland, Oregon USA \emph{PREPRINT}}
\CopyrightYear{2010} % Allows default copyright year (200X) to be over-ridden - IF NEED BE.
%\crdata{0-12345-67-8/90/01}  % Allows default copyright data (0-89791-88-6/97/05) to be over-ridden - IF NEED BE.
% --- End of Author Metadata ---

\title{Superior Exploration-Exploitation Balance with Quantum-Inspired Hadamard Walks}
%\subtitle{Late-Breaking Abstract\titlenote{We hope to communicate further analysis in this direction in the future.}}
%subtitle{[Extended Abstract]
%\titlenote{A full version of this paper is available as
%textit{Author's Guide to Preparing ACM SIG Proceedings Using
%\LaTeX$2_\epsilon$\ and BibTeX} at

\numberofauthors{2} %  in this sample file, there are a *total*
\author{
% 1st. author
\alignauthor
Sisir Koppaka\\
       \affaddr{Depts. of Mechanical and Industrial Engg.}\\
       \affaddr{IIT Kharagpur}\\
       \affaddr{Kharagpur, WB 721302, India}\\
       \email{sisir.koppaka@acm.org}
% 2nd. author
\alignauthor
Ashish Ranjan Hota\\
       \affaddr{Dept. of Electrical Engg.}\\
       \affaddr{IIT Kharagpur}\\
       \affaddr{Kharagpur, WB 721302, India}\\
       \email{hota.ashish@acm.org}
}
% There's nothing stopping you putting the seventh, eighth, etc.
% author on the opening page (as the 'third row') but we ask,
% for aesthetic reasons that you place these 'additional authors'
% in the \additional authors block, viz.
\additionalauthors{}
\date{30 July 1999}
% Just remember to make sure that the TOTAL number of authors
% is the number that will appear on the first page PLUS the
% number that will appear in the \additionalauthors section.

\maketitle
\begin{abstract}
This paper extends the analogies employed in the development of quantum-inspired evolutionary algorithms by proposing quantum-inspired Hadamard walks, called QHW. A novel quantum-inspired evolutionary algorithm, called HQEA, for solving combinatorial optimization problems, is also proposed. The novelty of HQEA lies in it's incorporation of QHW Remote Search and QHW Local Search - the quantum equivalents of classical mutation and local search, that this paper defines. The intuitive reasoning behind this approach, and the exploration-exploitation balance thus occurring is explained. From the results of the experiments carried out on the 0,1-knapsack problem, HQEA performs significantly better than a conventional genetic algorithm, CGA, and two quantum-inspired evolutionary algorithms - QEA and NQEA, in terms of convergence speed and accuracy.
\end{abstract}
\category{I.2.8}{Artificial Intelligence}{Problem Solving, Control Methods, and Search}
%A category including the fourth, optional field follows...
%\category{D.2.8}{Software Engineering}{Metrics}[complexity measures, performance measures]

\terms{Algorithms, Experimentation, Theory}

\keywords{Evolutionary Algorithms, Quantum Computing, Random Walks, Combinatorial Optimization, Knapsack Problem, Late Breaking Abstract}

\section{Introduction}
In recent years, several quantum-inspired evolutionary algorithms have been proposed, drawing upon quantum computing concepts such as quantum bits and the superposition of states. Q-bit representations for individuals and quantum gate-based evolution operators have been proposed. It was experimentally observed that the performance of quantum-inspired class of evolutionary algorithms was superior to that of conventional genetic algorithms.\cite{Han2002,Mahdabi2009}. 

In conventional genetic algorithms, the issues of exploration and exploitation were balanced through various mutation methods, and local search techniques. However, the equivalents of mutation and local search are unexplored territories in quantum-inspired evolutionary algorithms.

 In this paper, the use of quantum-inspired Hadamard walks for both exploration and exploitation through QHW Remote and Local Search respectively is demonstrated.

\section{QHW}

\begin{figure}[h]
\centering
\epsfig{file=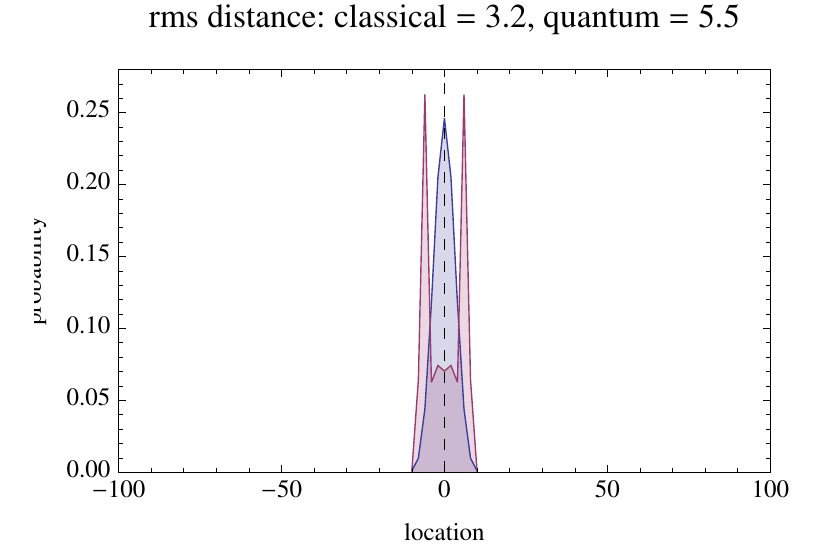, scale=0.75}
\caption{QHW for  $n=10, n_{max}=100$}
%\caption{A sample black and white graphic (.eps format).}
\end{figure}

\begin{figure}[h]
\centering
\epsfig{file=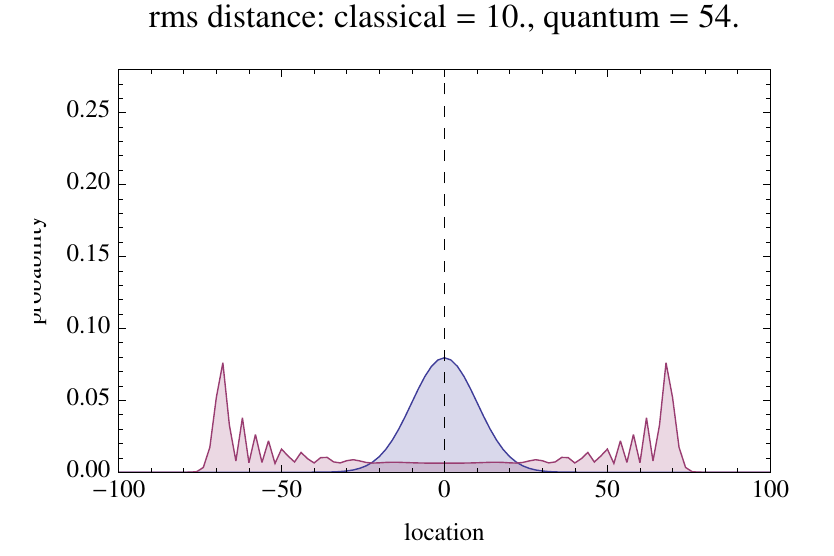,scale=0.75}
\caption{QHW for $n=100, n_{max}=100$}
%\caption{A sample black and white graphic (.eps format).}
\end{figure}
Quantum walks have been known for showing properties quite unlike classical random walks\cite{Kempe2008}. A frequently used balanced unitary coin, which lends additional degrees of freedom to the quantum walk over it's classical counterpart, is the Hadamard coin \textit{H},

\begin{center}{\textit{H}=$\frac{1}{\sqrt{2}} \left( \begin{array}{cc} 1 & i \\ i & 1 \end{array} \right)$}\end{center}

\begin{algorithm}[h]
\caption{QHW}
\label{QHW}
\floatname{algorithm}{Procedure}
\begin{algorithmic}
\STATE Initialize the Hadamard Walk centered at $\Delta \theta=0$ with $2n+1$ states from $[-n,n]$
\STATE Map $[-n_{max},n_{max}]$ to $[-\pi,\pi]$
\STATE Initialize the quantum state as $\psi_{00}=\frac{i}{\sqrt{2}}$ ; $\psi_{10}=\frac{1}{\sqrt{2}}$
\STATE Do the Hadamard walk by flipping the \textit{Hadamard coin} n times
\STATE Probability of $\Delta \theta_{k}$ (state $k$) is given by ${|\psi_{0,k}|}^{2}+{|\psi_{1,k}|}^{2}$
\STATE Determine $\Delta \theta$ given the weighted probability distribution of $\Delta \theta_{k}$
\RETURN  $\Delta \theta$ 
\end{algorithmic}
\end{algorithm}

Superposition concedes interesting properties to the quantum walk. The quantum walk propagates relatively faster along the line: it's variance grows \textit{quadratically} with the number of steps \textit{n}, $\sigma^2\propto{n^2}$, compared to $\sigma^2\propto{n}$ for the classical random walk. QHW takes advantage of this rapid convergence as shown in \textbf{Figure 1-2}, and \textbf{Algorithm 1}.

\section{QHW Local and Remote Search}
The novelty of QHW lies in that it can be used for both remote searches as well as local searches, by suitably altering the parameter $n$ relative to $n_{max}$. By operating with a high $n$ value on the worst solutions, \textit{exploration} is achieved. By using a smaller $n$ value on the good solutions, \textit{exploitation} is achieved. 

The algorithm for QHW Local and Remote Search is surmised below as \textbf{Algorithm 2}.
\begin{algorithm}
\caption{QHW Local and Remote Search}
\label{QHW Local and Remote Search}
\floatname{algorithm}{Procedure}
\begin{algorithmic}
\STATE Select $best$ (for local) and $worst$ (for remote)$\ n\%$ of population by fitness values
\STATE $i \leftarrow 0$
\FOR {each of the selected $best$/$worst\ n\%$ individuals}
\WHILE{$i < m$}
\STATE $\theta'=\theta+\Delta \theta$ where $\Delta \theta=$ \textbf{QHW}[$n$]
\STATE Generate new $\alpha_{i}',\beta_{i}'$ by using $\theta'$
\IF{(Fitness[NewIndividual]>Fitness[OriginalIndividual])}
\STATE Replace OriginalIndividual with NewIndividual
\ENDIF
\ENDWHILE
\ENDFOR
\RETURN Updated $n\%$ individuals
\end{algorithmic}
\end{algorithm}

\section{HQEA}
\textbf{Algorithm 3} describes a novel quantum-inspired evolutionary algorithm that utilizes QHW Remote Search for exploration and QHW Local Search for exploitation.
\begin{algorithm}
\caption{HQEA}
\label{HQEA}
\floatname{algorithm}{Procedure}
\begin{algorithmic}
\STATE $t \leftarrow 0$
\STATE Initialize Population $Q(t)$, Make $P(t)$, Evaluate $P(t)$, Store $P(t) \rightarrow B(t)$
\WHILE{$t<t_{max}$}
\STATE $t \leftarrow t+1$
\STATE Make $P(t)$ by observing states of $Q(t)$
\STATE Repair $P(t)$
\STATE Evaluate $P(t)$
\STATE do \textbf{QHW Remote Search}
\STATE do \textbf{QHW Local Search}
\STATE Update $Q(t)$
\STATE Store best solutions among $B(t-1), P(t) \rightarrow B(t)$
\STATE Store best solution $b \leftarrow B(t)$
\IF{($migration\ period$)}
\STATE migrate $b$ or $b^{t}_{j}$ to $B(t)$ $locally$ or $globally$
\ENDIF
\STATE display $b(t)$
\ENDWHILE
\end{algorithmic}
\end{algorithm}

\section{Experiments}
The performance of HQEA to that of a conventional genetic algorithm, CGA, and two quantum-inspired evolutionary algorithms, QEA and NQEA, on the 0-1 knapsack problem is compared.

Table 1 reports the average fitness of the best solutions found by the respective algorithms over 10 runs on two instances of the 0-1 knapsack problem with 200 and 500 items, rounded to the nearest integer. In the experiments of HQEA, the parameters are set to $n=10$ for QHW Local Search and $n=100$ for QHW Remote Search, with $n_{max}=100$ for both. All other parameters for HQEA and those of QEA and NQEA are set according to those proposed  in  \cite{Han2002} and \cite{Mahdabi2009}, with a population of 10 individuals and $\Delta \theta=0.01\pi$. In CGA, a uniform crossover, and roulette wheel selection are used. 

\begin{table}[h]
\centering
\caption{Results on 0-1 knapsack problem}
\begin{tabular}{|c| c| c| c| c| c|}\hline
Items Size&Iterations&CGA&QEA&NQEA&HQEA\\ \hline
 & 50 & 943 & 956 & 966 & 971\\
& 100 & 946 & 965 & 971 & 984\\
 200& 250 & 949 & 995 & 1002 & 1009\\
 & 500 & 955 & 1015 & 1018 & 1031\\
 & 1000 & 958 & 1025 & 1030 & 1032\\ \hline
& 50 & 2647 & 2656 & 2661 & 2675\\
& 100 & 2656 & 2667 & 2671 & 2677\\
500 & 250 & 2660 & 2672 & 2677 & 2684\\
 & 500 & 2674 & 2676 & 2681 & 2689\\
 & 1000 & 2678 & 2689 & 2693 & 2702\\ \hline
\end{tabular}
\end{table}

\section{Conclusions}
In this paper, the convergence and accuracy in HQEA is noted to be significantly quicker compared to CGA, QEA and NQEA. Note that HQEA does not build upon NQEA - rather, it is a new and innovative look at quantum-inspired evolutionary algorithms.

%\end{document}  % This is where a 'short' article might terminate

\bibliographystyle{abbrv}
\bibliography{qea}  % sigproc.bib is the name of the Bibliography in this case
% ACM needs 'a single self-contained file'!
%\balancecolumns
%Appendix A

%\balancecolumns % GM June 2007
% That's all folks!
\end{document}